\documentclass[journal]{IEEEtran}

\ifCLASSINFOpdf
\else
\fi

\usepackage{mathrsfs}

\usepackage{graphicx}
\usepackage{amsmath}
\usepackage{multirow}
\usepackage{mathtools}
\usepackage{placeins}
\usepackage{float}

\usepackage{booktabs}
\usepackage{ragged2e}
\usepackage{algorithm}
\usepackage{algorithmic}
\usepackage{amsmath}
\usepackage{graphics}
\usepackage{epsfig}
\usepackage{amssymb}
\usepackage{threeparttable}

\usepackage{bm}
\usepackage{subcaption}
\usepackage{rotating}
\usepackage{hyperref}
\usepackage{cite}

\usepackage{stackengine}
\usepackage{stmaryrd}

\usepackage{footnote}
\renewcommand{\algorithmicrequire}{\textbf{Input:}}
\renewcommand{\algorithmicensure}{\textbf{Output:}}

\begin{document}

\title{Weak Disambiguation for Partial Structured Output Learning  }

\author{Xiaolei~Lu,Tommy~W.S.Chow,~\IEEEmembership{Fellow,~IEEE}%

\IEEEcompsocitemizethanks{\IEEEcompsocthanksitem Xiaolei Lu is with
the Dept of Electronic Engineering at the City University of Hong
Kong, Hong Kong (Email:xiaoleilu2-c@my.cityu.edu.hk)
\IEEEcompsocthanksitem Tommy W S Chow is with the Dept of Electronic
Engineering at the City University of Hong Kong, Hong Kong (E-mail:
eetchow@cityu.edu.hk).}}

\maketitle

\begin{abstract}

Existing disambiguation strategies for partial structured output learning just cannot generalize well to solve the problem that there are some candidates which can be false positive or similar to the ground-truth label. In this paper, we propose a novel weak disambiguation for partial structured output learning (WD-PSL). First, a piecewise large margin formulation is generalized to partial structured output learning, which effectively avoids handling large number of candidate structured outputs for complex structures. Second, in the proposed weak disambiguation strategy, each candidate label is assigned with a confidence value indicating how likely it is the true label, which aims to reduce the negative effects of wrong ground-truth label assignment in the learning process. Then two large margins are formulated to combine two types of constraints which are the disambiguation between candidates and non-candidates, and the weak disambiguation for candidates. In the framework of alternating optimization, a new 2n-slack variables cutting plane algorithm is developed to accelerate each iteration of optimization. The experimental results on several sequence labeling tasks of Natural Language Processing show the effectiveness of the proposed model.

\end{abstract}


\IEEEpeerreviewmaketitle

\section{Introduction}
%
%
%
%
\IEEEPARstart{S}{tructured} output learning, which refers to learn a mapping function from both structured input and output (e.g., sequence and tree), has been widely used in Natural Language Processing and Computational Biology. For example, structured output learning in Part-of-Speech (POS) tagging is to predict the POS of each word for the sentence sequence and produce a POS sequence of equal length. In gene prediction, structured output learning aims to find genes in a geometric DNA sequence.

Many effective methods like probabilistic graph models and structured SVMs (S-SVM) \cite{re1} have been proposed and have delivered promising results to structure prediction. Learning a generative or discriminative classifier requires large number of training sequences with fully annotations, which is costly and laborious to produce. As a result, semi-supervised learning methods have been proposed to alleviate the burden of manual annotation, but they still need exact annotations for structured outputs. In real-life applications, we are often given with large number of partially (or ambiguously) annotated structured outputs.

 For example, as shown in Figure 1(a), it is more efficient to annotate certain parts of a sentence. Besides, Figure 1(b) shows the ambiguous annotations. Since annotators may not be specialists in linguistics, a word in the sentence can have multiple POS tags. Therefore, incorporating partial annotations into structured out learning are practically significant.

\begin{figure}
\centering
\includegraphics[width=3.5in,height = 2.5in ]{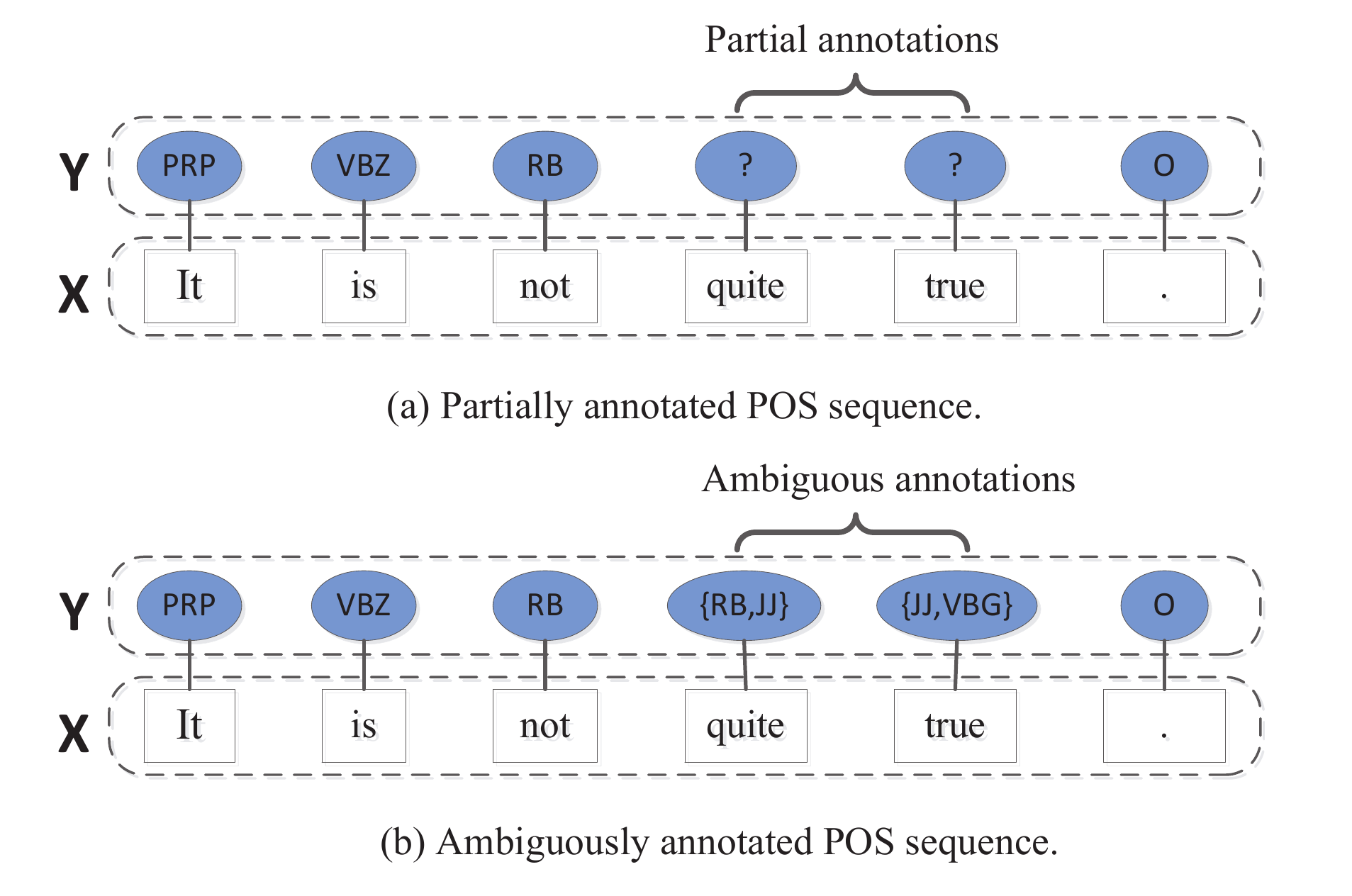}
\caption {Examples of partially (or ambiguously) annotated POS
sequences.} \label{fig:secondfigure}
\end{figure}

In recent years a weakly-supervised learning framework, partial-label learning (PLL), has been proposed to solve the problem of ambiguous annotations, which arises from real world scenarios \cite{re2}. For example, given an image, annotators with different knowledge probably label it differently, as illustrated in Figure 2. In partial-label learning, the ground-truth label is masked by ambiguous annotations. The datasets with large proportion (or full) of ambiguous annotations are of poor label quality. Therefore how to identify the ground-truth label from ambiguous annotations is more important for PLL-based methods. The objective of PLL is to learn a model from ambiguously labeled instances. PLL recovers the correct label from ambiguous labels and then predicts unseen examples \cite{re2}. 

\begin{figure}[H]
\centering
\includegraphics[width=3.5in,height = 1.0in ]{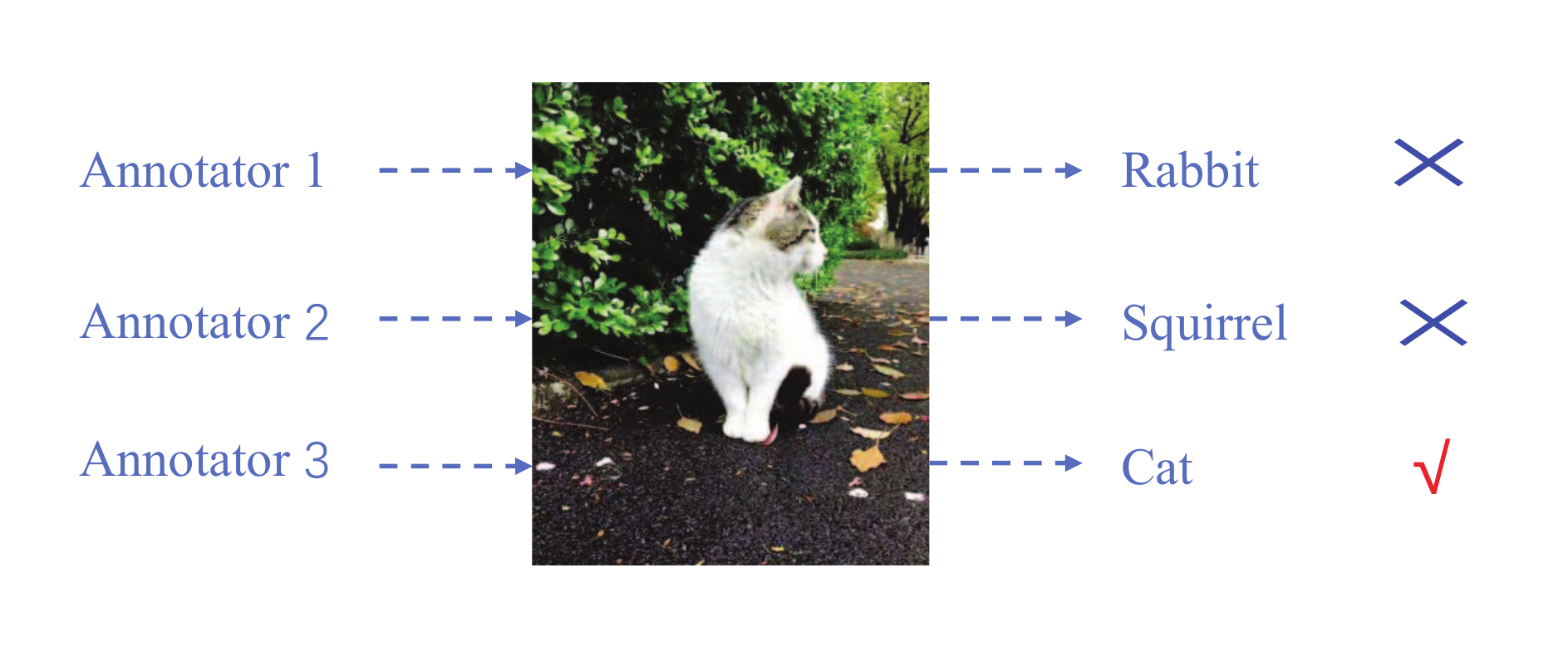}
\caption {An example of partial label learning.}
\label{fig:secondfigure}
\end{figure}

Extending the setting of partial label learning to structured output learning (i.e., partial structured out learning), some of the structured outputs are annotated with ground-truth labels while the other elements are ambiguously annotated. A straightforward way for partial structured out learning is to enumerate all possible structured outputs and treat them as the ground-truth. Given an ambiguously annotated sequence with length $L$, each element is assigned with $s$ candidate labels, then the number of possible structured outputs is $s^{L}$, which is computationally expensive in real-word applications. For example, POS tagging for long sentences. There are certain research work focusing on combining PLL methods with sequence labeling, which aims to identify the ground-truth label sequence from exponential number of candidates. However, in many scenarios there exist some candidates that are very difficult to disambiguate with the ground-truth label sequence. In iterative optimization of identification disambiguation strategy, the score of the correct structured output can be lower than (or equal to) that of some candidate structured outputs. For example, as shown in Figure 1(b), the ground-truth is ``PRP/VBZ/RB/RB/JJ/O" while ``PRP/VBZ/RB/JJ/JJ/O" is more likely to be assigned as the true label sequence, which will adversely affect the parameter learning for next iteration.

Traditional structured output learning methods (e.g., CRFs and HMMs) require repeated inference, which is intractable for complex structure (e.g. lattice). Sutton and McCallum \cite{re3} proposed piecewise training to decompose complex structures into tractable subgraphs that are called ``pieces", and perform inference on small subgraphs. Alahari et al. \cite{re4} extended piecewise training to large margin formulation for CRFs, in which the random field is decomposed into pieces which can be treated as individual training samples. Piecewise training for structured output learning models with maximum margin criteria can be more computationally efficient as it does not need to perform inference. 
 
In this paper, to address the above two problems in learning from partial annotations, we propose weak disambiguation for partial structured output learning (WD-PSL). By addressing different contribution of each candidate in parameter learning, weak disambiguation strategy avoids identifying the unique ground-truth label sequence and thus benefits the learning from candidates that are very difficult to disambiguate with the ground-truth label, which greatly reduces the negative effect of wrong ground-truth label assignment in optimization. Our contributions can be summarized as follows:

First, by generalizing the piecewise large margin formulation to partial annotations, WD-PSL dose not need to perform inference compared with traditional graph models for sequence labeling (e.g., HMMs and CRFs). Moreover, piecewise training for WD-PSL effectively avoids handling large number of candidate structured outputs for complex structures, which is more computationally efficient for parameters learning.

Second, two large margins are formulated to combine two types of constraints which are the disambiguation between candidates and non-candidates, and the weak disambiguation for candidates. Further, a new 2n-slack variables cutting plane algorithm is developed to accelerate the iteration of alternating optimization. In weak disambiguation strategy, each candidate label is assigned with a confidence value to indicate how likely it is the ground-truth. By addressing different contribution of the label in candidates, WD-PSL greatly reduce the negative effects of wrong ground-truth label assignment in the learning process, which can help improve the performance of sequence labeling.

Third, in order to demonstrate the effectiveness of the proposed model, we evaluate PLL-based structured output learning methods on several NLP sequence labeling tasks. The experimental results show that our proposed model outperforms the traditional methods.

\section{Related Work}

Probabilistic graph models can effectively capture dependency relationship between structured outputs. Hidden Markov Models (HMMs) \cite{re5}\cite{re6} and Conditional Random Fields (CRFs) \cite{re7}\cite{re8}\cite{re9} are the most popular graph models in structured output learning. Many variants of basic graph models have been proposed to predict complex structured outputs. For example, Sarawagi and Cohen \cite{re10} proposed semi-Markov conditional random fields (semi-CRFs) to segment structured outputs, which is very useful in the tasks of chunking and named entity recognition. Furthermore, hierarchical conditional random fields \cite{re11} was designed to model multiple structured outputs. In recent years, CRF-CNN \cite{re12} and LSTM-CRF \cite{re13}, which arise from the combination of deep learning and graph models, can achieve competitive results compared with traditional graph models.

Generally, existing methods for structured output learning require large number of training instances with ground-truth annotations, which are challenging conditions. First, the annotators should be equipped with strong domain knowledge base to obtain precisely labelings, which is financially expensive for labour costs. Further, it is prohibitively time-consuming to manually label complex structured outputs than traditional discrete values. For example, label assignment of document classification is in the document-level while in POS tagging we need to annotate each word in all sentences level.

Semi-supervised learning has been introduced to structured output learning as it can effectively incorporate large number of unannotated instances in the optimization. Jiao et al. \cite{re14} proposed semi-supervised CRFs to improve sequence segmentation and labeling. By minimizing the conditional entropy on unlabeled training instances and then combining with the objective of CRFs, semi-supervised CRFs achieve improved performance. Brefeld and Scheffer \cite{re15} developed co-training principle into support vector machine to minimize the number of errors for labeled data and the disagreement for the unlabeled data, which can outperform fully-supervised SVM in specific tasks. Although semi-supervised based structured output learning models partly reduce the number of annotated instances, these instances still need exact labelings.

Partial-label learning (PLL) has been proposed to solve the problem of ambiguous annotations. Most of PLL algorithms focus on two disambiguation strategies: ``Average Disambiguation (AD)'' and ``Identification Disambiguation (ID)" \cite{re16}. AD strategy treats each candidate label equally by averaging the scores of all candidate labels. Cour et al. \cite{re2} proposed Convex Loss for Partial Labels (CLPL) to disambiguate candidate labels with non-candidate labels. Another instance-based AD solution is to construct a nonparametric classifier and keep ambiguous label information. For example, the ground-truth label can be predicted by k-nearest neighbors weighted voting \cite{re17}. ID strategy, which aims to identify the ground-truth label from candidate labels, has been widely adopted as AD-based methods incorporate wrong label information by combining the outputs of all candidate labels. ID-based discriminative graph models treat the ground-truth label as a latent variable and identify it by the iterative maximum likelihood method. Jin et al. \cite{re18} proposed ``EM+Prior" model by generalizing the EM model with prior knowledge indicating which label is more likely to be the ground-truth label. Moreover, maximum margin learning provides an effective way to disambiguate ground-truth labels with ambiguous labels. Partial label SVM (PL-SVM) \cite{re19} was formulated to maximize the margin between current prediction of the ground-truth label and the best wrong prediction of non-candidate labels. Yu and Zhang \cite{re20} proposed Maximum Margin Partial Label Learning to address the predictive difference between the ground-truth label and other candidate labels. Further, Zhou et al. \cite{re21} extended Gaussian process model to partial label learning to deal with nonlinear classification problems. For ID-based strategy, however, in most cases ground-truth labels do not have the maximum score in the iterative label assignment, which inevitably degrades the performance of the subsequent classification. More recently, Zhang et al. \cite{re22} proposed a disambiguation-free strategy, which treats candidate label set as an entirety by using Error-Correcting Output Codes (ECOC) coding matrix. Although this strategy is simple and effective, generating binary training set from partially labeled data relies on coding matrix. 

Partial structured output learning combines the setting of partial label learning into sequence labeling, where some of the structured outputs are annotated with ground-truth labels while the other elements are ambiguously annotated, as shown in Figure 1(b). A simple strategy is to treat all possible label sequences as the ground-truth. Tsuboi et.al \cite{re23} proposed a marginalized likelihood for CRFs, which enumerates all label sequences that are consistent with ambiguous annotations. Since it is not practical to evaluate all label configurations as the number of possible label sequences is exponential to the number of ambiguous annotated elements, constrained lattice training \cite{re24}\cite{re25} was proposed to disallow invalid label sequences. More recently, there are some research work generalizing partial label learning to structured output learning. Lou and Hamprecht \cite{re26} used large margin based methods to discriminate the ground-truth structured output from other possible structured outputs and proposed accelerated concave-convex procedure for iterative optimization. Li et al. \cite{re27} further decomposed a large margin formulation into two parts: maximize the margin between the ground-truth and non-candidates; maximize the margin between the ground-truth and other candidates. However, these methods for partially structured output learning have to discriminate the ground-truth from large number of possible structured outputs, which is inefficiently in the optimization. Furthermore, as mentioned in the above PLL research work, unique ground-truth identification strategy may results in wrong label assignment as there exist some candidates that are very difficult to disambiguate with the ground-truth label sequence.

\section{Preliminaries}
In this section, we briefly introduce two learning methods related to our work, namely partial label learning and piecewise large margin learning.

\subsection{Partial Label Learning}
Given training samples $D=\left \{ x_{i},S_{i} \right \}_{i=1}^{n}$, $S_{i}\in \mathcal{Y}$ is the set of candidate labels, $\mathcal{Y}$ is the output space for all possible labels. Partial label learning aims to learn a model from ambiguously labeled samples and then generalize to unseen examples. Existing PLL methods mainly focus on Average Disambiguation (AD) Strategy and Identification Disambiguation (ID) Strategy.

AD strategy treats each candidate label equally by averaging the outputs of all candidate labels. The learning formulation is based on one-against-all scheme which treats the data $\left \{ x,y \right \}$ that $y\in \mathcal{Y}\backslash S_{i}$ as negative training samples. The resulting loss called Convex Loss for Partial Labels (CLPL) is defined as

\begin{equation}
\scalebox{0.8}{$ L^{CLPL}(\omega )=\psi \left ( \frac{1}{\left |
S_{i} \right |}\sum_{y\in S_{i}}F(x_{i},y) \right )+\sum_{y\in
\mathcal{Y}\backslash S_{i}}\psi\left ( -F\left ( x_{i} ,y\right )
\right ),$}
\end{equation}
where $\psi \left ( \cdot  \right )$ denotes an upper bound on the $0\slash1$ loss. $F(x_{i},y)$ is the score of label $y$ given input $x_{i}$. If $S_{i} $ contains only one label, CLPL transforms to a general multiclass loss.

Apart from disambiguation between candidate labels and non-candidate labels, ID strategy adopts pairwise comparison scheme which further addresses the disambiguation between the ground-truth label and other candidate labels. By differentiating the current prediction  (the maximum output from from candidate labels) and the best wrong prediction (maximum output from non-candidate labels), the learning formulation is expressed as 

\begin{equation}
\scalebox{0.8}{$ L^{MMPL}(\omega )=\psi \left ( \max\limits_{y\in
S_{i}}F(x_{i},y)-\max\limits_{y\in \mathcal{Y}\backslash
S_{i}}F\left ( x_{i} ,y\right )\right ),$}
\end{equation}
where $\psi \left ( \cdot  \right )$ represents the hinge loss. $F(x_{i},y)$ is the score of label $y$ given input $x_{i}$.

Candidate labels contain much wrong label information, which obviously contribute differently to the learning process. Although ID strategy can alleviate the shortcoming of AD based approaches, in most cases some candidate labels are very difficult to disambiguate with the ground-truth label. Incorrect label assignment in iterative optimization will degrade the performance of classifier.

\subsection{Piecewise Large Margin Learning}
The combination of piecewise training and max-margin learning can learn a discriminative classifier efficiently. Each vertex $i$, as shown in Figure 3(b), represents a tree-structured graph. Piecewise training can treat a single factor (denoted with black square in Figure 3(b) as a ``piece" of the graph. The energy function on this tree-structured graph for a vertex $i$ is defined as

\begin{equation}
E^{i}(y)=\theta ^{T}f(i,j,x,y)+b
\end{equation}
where $i$ is the set of all nodes in the tree-structured graph. $j=\left \{ j|j\in N_{i} \right \}$, $N_{i}$ denotes the neighbors of vertex
$i$. $f(i,j,x,y)$ is the feature which combines unary and pairwise features. $b$ is the bias value.

\begin{figure}[H]
\centering
\includegraphics[width=3.5in,height = 1.0in ]{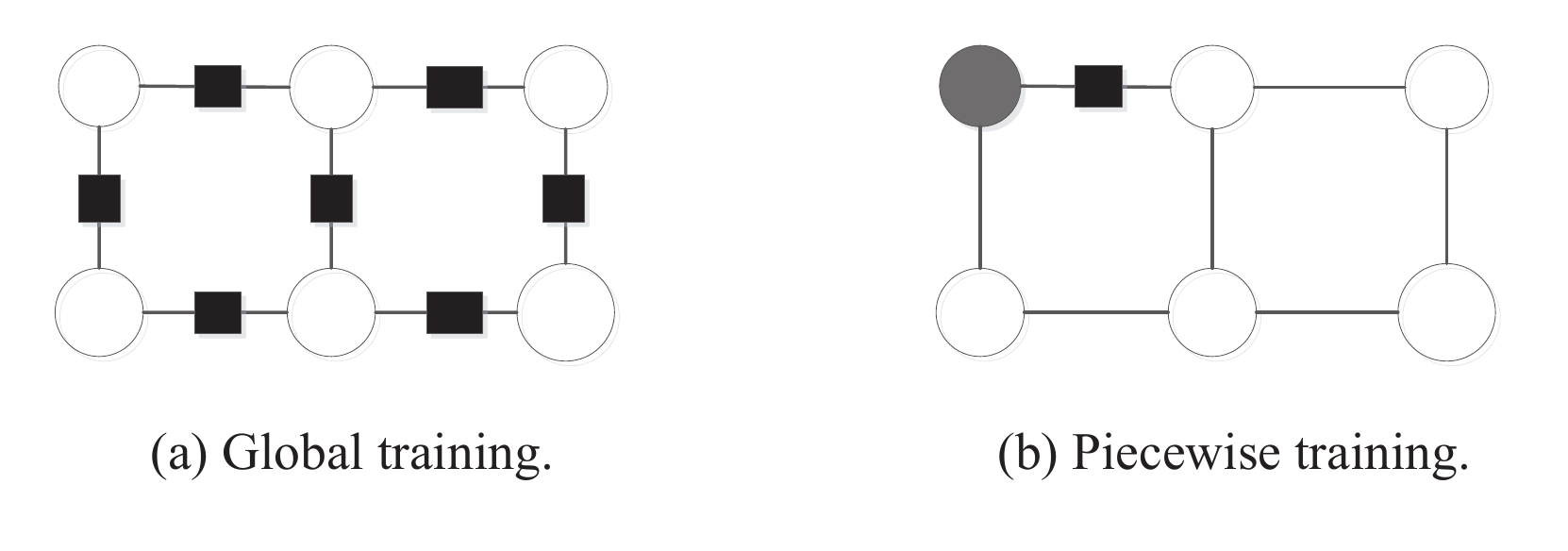}
\caption {Global training and piecewise training.}
\label{fig:secondfigure}
\end{figure}

Let $M_{+}$ and $M_{-}$ denote the number of positive and negative training samples respectively. The parameters can be learned by solving the following soft-margin optimization problem:

\begin{equation}
(\theta ^{*},b^{*}) = \arg\min\frac{1}{2} \left \| \theta  \right
\|^{2}+ C(\sum_{m}\xi _{+}^{m}+\sum_{n}\xi _{-}^{n}),
\end{equation}

\begin{equation}
s.t.  \quad \theta ^{T}f_{+}^{m}(i,j,x,y) + b\geq 1-\xi
_{+}^{m},\forall m,
\end{equation}
\begin{equation}
\qquad\theta ^{T}f_{-}^{n}(i,j,x,y) + b\leq  -1+\xi _{-}^{n},\forall
n,
\end{equation}
\begin{equation}
\qquad  \xi _{+}^{m}\geq 0,\forall m\in \left \{ 1,...,M_{+} \right
\},
\end{equation}
\begin{equation}
\qquad  \xi _{-}^{n}\geq 0,\forall n\in \left \{ 1,...,M_{-} \right
\}.
\end{equation}
where $C$ is the regularization parameter which controls the trade-off between the misclassification on the training data and the width of margin. $\xi$ is the slack variable that is introduced to allow non-linearly seperable case.

Training procedure in piecewise large margin learning does not need to perform inference, which is more computationally efficient than traditional graph-based structured output learning methods. The following section will discuss how to extend piecewise large margin learning to partial annotations.

\section{WD-PSL algorithm}

In this section, we firstly introduce how to combine piecewise large margin learning with partial label learning for partially annotated structured outputs, then we present the formulation of WD-PSL algorithm and the efficient optimization method for parameters learning. Also, a comparative analysis between the proposed objective function and other partial losses is described.

\subsection{Formulation}

Given training sequences $\left \{ X^{i},Y^{i} \right \}_{i=1}^{N}$, $X^{i}=\left \{ x_{1},...,x_{m} \right \}$, $Y^{i}=\left \{ \textbf{y}_{1},...,\textbf{y}_{m} \right \}$, $\textbf{y}_{m}=\left\{ y_{1},...,y_{l} \right \}$ which denotes the set of candidate labels. There are $l^{m}$ candidate structured outputs for each input sequence. Linear-chain CRFs model the structured outputs in a way shown in Figure 4(a), where the black square is the transition factor. To reduce the large number of candidate structured outputs, the random fields are divided into pieces which are presented in Figure 4(b). For example, $X^{i}$ can be divided into pieces with two transition factors. Then the total number of candidate structured outputs for $i_
{th}$ sequence is $(m-2)*l^3$.

Piecewise large margin learning treats each piece as independent training sample. Given training pieces $\left \{ x^{i},\textbf{y}^{i} \right \}_{i=1}^{n}$, where $\textbf{y}^{i} = \left \{ y_{1},..,y_{s} \right \}$ is the set of candidate structured outputs and $\textbf{y}^{i}$ contains the ground-truth structured output. $\mathcal{Y}$ denotes all possible structured outputs. The energy function in Equation (3) is generalized to the setting of partial annotations as follows:

\begin{equation}
E(x^{i},\textbf{y}^{i})=\sum_{y\in \textbf{y}^{i}}\sum_{k}\left ( \omega
^{T}f(j,k,x^{i},y) +b\right ),
\end{equation}
where $k$ denotes the set of nodes in $x^{i}$ and $j$ indexes the neighbors of $k$. $f(j,k,x^{i},y)$ is the feature vector concatenated by the state feature $f(x_{k}^{i},y_{k})$ and transition feature $f(y_{j},y_{k})$. $b$ is the bias value.

\begin{figure}[H]
\centering
\includegraphics[width=3.5in,height = 2.0in ]{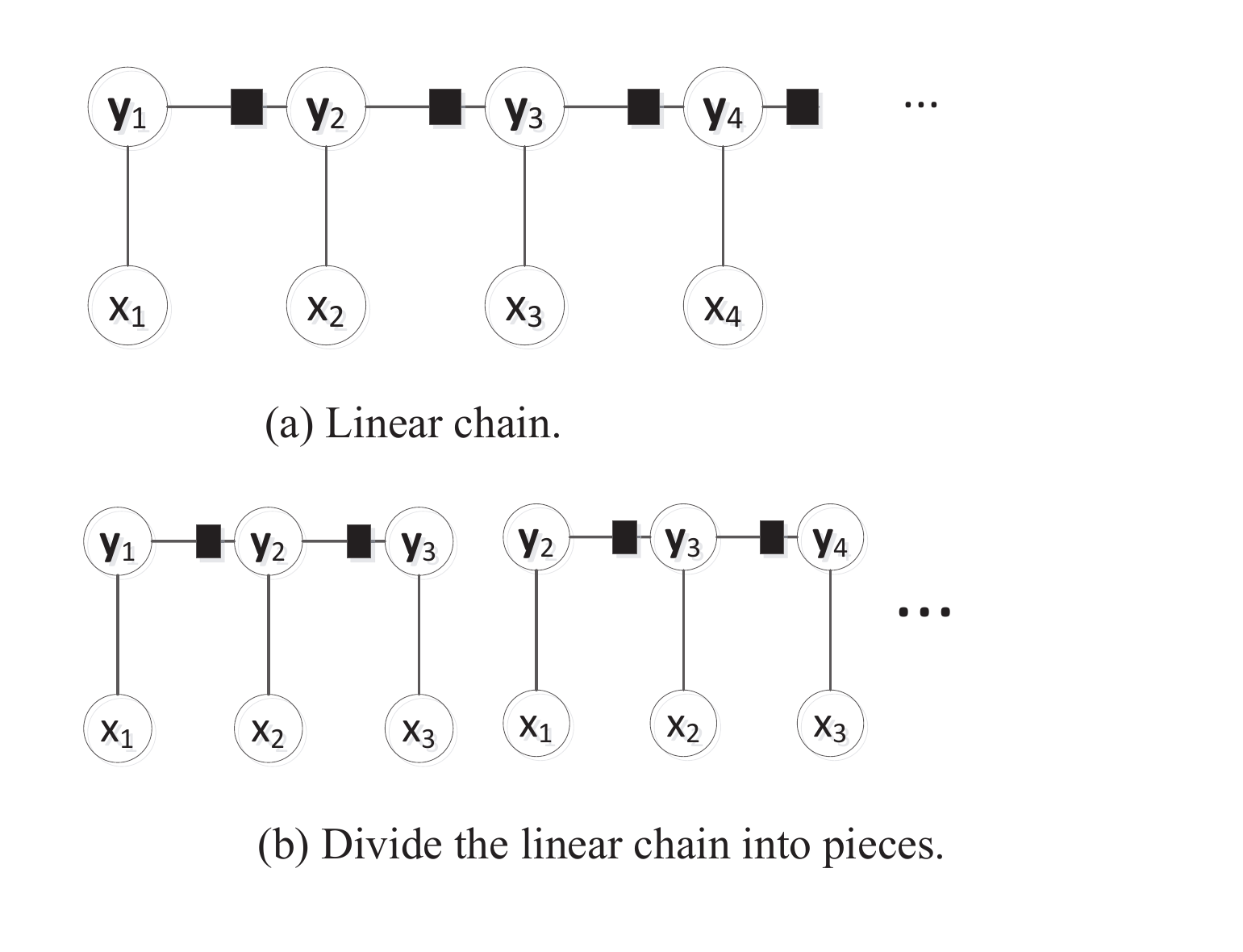}
\caption {Decomposing random fields into pieces.}
\label{fig:secondfigure}
\end{figure}

The proposed weak disambiguation strategy aims to disambiguate the candidates with non-candidates and reduce the negative effects of wrong ground-truth label assignment in the learning process, which is stated as follows:

\begin{figure}[H]
\centering
\includegraphics[width=4in,height = 2.5in ]{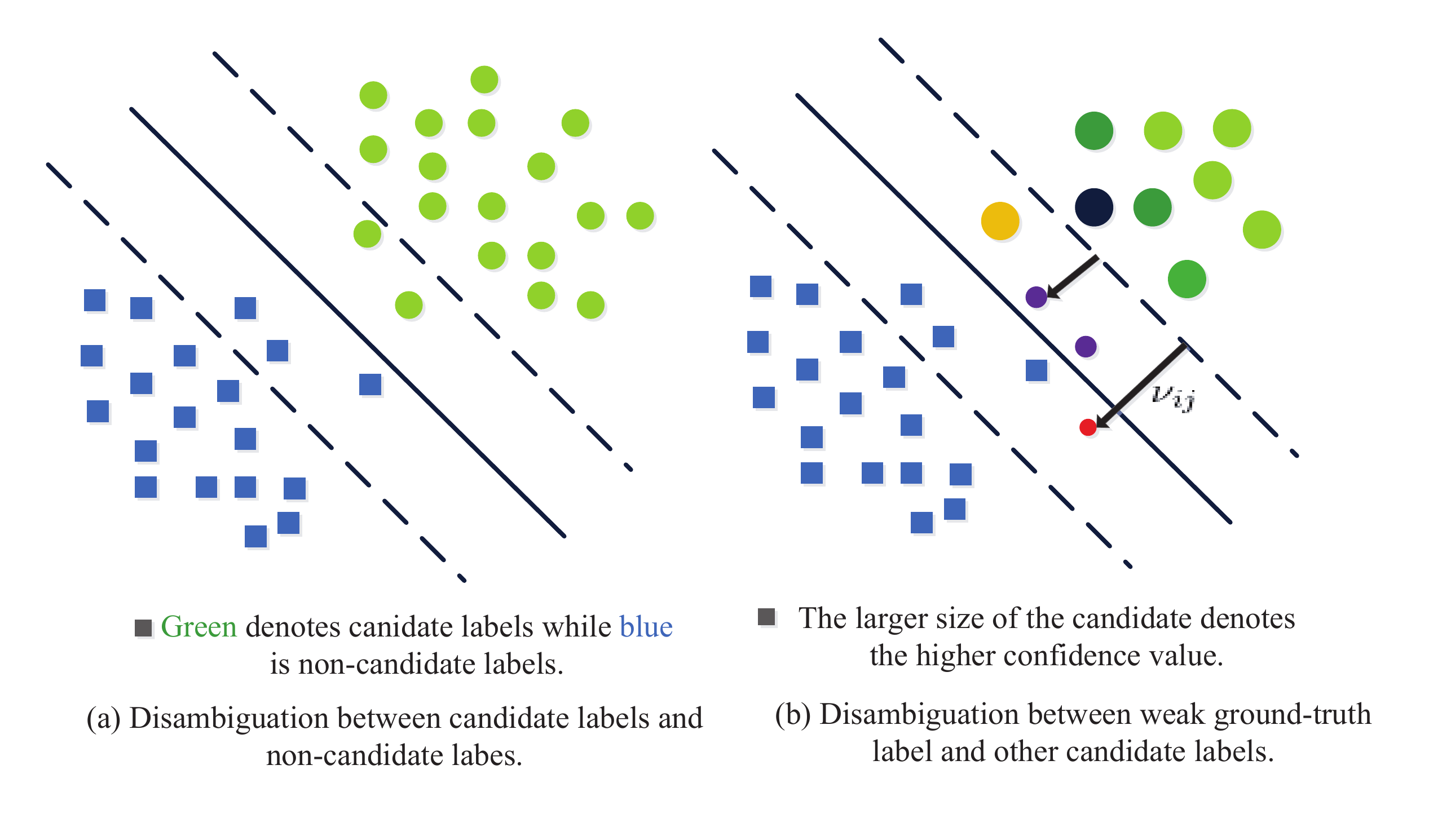}
\caption {Visualization of weak disambiguation strategy.}
\label{fig:secondfigure}
\end{figure}

First, disambiguation between candidate and non-candidate labels. As visualized in Figure 5(a), the weak disambiguation strategy employs the hard disambiguation for non-candidates, which can improve the discriminability of the model. A soft max-margin formulation is expressed as

\begin{equation}
\min \limits_{\xi } \frac{C_{1}}{n}\sum_{i=1}^{n}\xi
_{i}
\end{equation}

\begin{equation}
s.t. \qquad  \Delta (\textbf{y}^{i},{y}'')+\left \langle \omega,E\left (
x^{i},{y}''\right ) \right \rangle -  \left \langle \omega,E\left (
x^{i},\textbf{y}^{i}\right ) \right \rangle \leq \xi_{i},
\end{equation}
\begin{equation}
\forall i\in \left \{ 1,2,...,n \right \},\forall {y}''\in\mathcal{Y},\xi _{i}\geq 0.
\end{equation}
where $C_{1}$ is the regularization parameter and $\xi$ denotes slack variable.

For the loss function $\Delta (\textbf{y}^{i},{y}'')$, we would like the margin between multi-true labels and the best runner-up to scale linearly with the number of multi-true labels. Then $\Delta (\textbf{y}^{n},{y}'')$ is defined as 
 
\begin{equation}
\Delta (\textbf{y}^{i},{y}'')= \left\{
\begin{array}{lcl}
s & & {y}''\notin
\textbf{y}^{i},\\
0  & & {y}'' \in \textbf{y}^{i},
\end{array} \right.
\end{equation}
where $s$ is the number of candidate labels in $\textbf{y}^{i}$.

Second, weak disambiguation for the candidate labels. In weak disambiguation strategy, each candidate label is assigned with a confidence value to indicate how likely it is the ground-truth label, which not only avoids the wrong ground-truth label assignment in ID strategy but addresses the different contributions of candidate labels. As shown in Figure 5(b), 
an instance tagged with multiple candidate labels equals multiple data points with different colors. For those data points fall within the margin, the point with candidate label assigned with higher confidence value should be close to the margin on the correct side of the separating hyperplane. For an instance $x^{i}$, each candidate label $y_{j} \in \textbf{y}^{i}$ is assigned with a confidence value $P_{ij}$, a soft max-margin which weighted by the confidence is defined as

\begin{equation}
\min\limits_{\nu } \frac{C_{2}}{n}\sum_{i=1}^{n}\sum_{y_{j} \in \textbf{y}^{i}} P_{ij}\nu_{ij}
\end{equation}

\begin{equation}
s.t. \qquad \left \langle \omega,E\left (
x^{i},{y_{j}}\right ) \right \rangle \geqslant 1-\nu
_{ij},
\end{equation}
\begin{equation}
\forall i\in \left \{ 1,2,...,n \right \},\forall y_{j} \in \textbf{y}^{i}, \nu_{ij}\geq 0.
\end{equation}
where $C_{2}$ is the regularization parameter and $\nu$ is the slack variable.

The unified objective function is obtained by incorporating Equation (10) and Equation (14), which is defined as 
\begin{equation}
J(\omega )=\min\limits_{\omega ,\xi ,\nu }\frac{1}{2}\left \| \omega  \right\|^{2}+ \frac{C_{1}}{n}\sum_{i=1}^{n}\xi_{i}+\frac{C_{2}}{n}\sum_{i=1}^{n}\sum_{y_{j} \in \textbf{y}^{i}} P_{ij}\nu_{ij}\end{equation}

\begin{equation}
s.t. \qquad  \Delta (\textbf{y}^{i},{y}'')+\left \langle \omega,E\left (
x^{i},{y}''\right ) \right \rangle -  \left \langle \omega,E\left (
x^{i},\textbf{y}^{i}\right ) \right \rangle \leq \xi_{i},
\end{equation}
\begin{equation}
\forall i\in \left \{ 1,2,...,n \right \},\forall {y}''\in\mathcal{Y},\xi _{i}\geq 0.
\end{equation}
\begin{equation}
 \qquad \left \langle \omega,E\left (
x^{i},{y_{j}}\right ) \right \rangle \geqslant 1-\nu
_{ij},
\end{equation}
\begin{equation}
\forall i\in \left \{ 1,2,...,n \right \},\forall y_{j} \in \textbf{y}^{i}, \nu_{ij}\geq 0.
\end{equation}

\subsection{Optimization}

In the above discussion, the confidence value for each candidate structured output of the instance is a constant for modeling. The initialization of confidence value for each candidate of instance $x^{i}$ can be defined as $P_{ij}=\frac{1}{\left | \textbf{y}^{i} \right |} $, where $j$ indexes the candidate structured output and $\left | \textbf{y}^{i} \right |$ is the number of candidate structured outputs for $x^{i}$. The optimization for Equation (17) can be solved by alternating optimization:

1. Fixed $\omega$. For each candidate structured output $y_{j}$ of instance $x^{i}$, the corresponding $P_{ij}$ is defined as
\begin{equation}
P_{ij}=\frac{E_{ij}-\min\left \{ E_{ij} \right \}_{j=1}^{s}}{\max\left \{ E_{ij} \right \}_{j=1}^{s}-\min\left \{ E_{ij} \right \}_{j=1}^{s}}
\end{equation}
where $s$ is the number of candidate structured outputs in $ \textbf{y}^{i} $.

2. Fixed $P$. Optimizing $\omega$ by solving the minimization of the Equation (17).

The parameters $\omega$ and $P$ can be learned by iterating the above two steps. The procedure of alternating optimization for WD-PSL is shown in
Algorithm 1.

\begin{algorithm}
\renewcommand{\algorithmicrequire}{\textbf{Input:}}
\renewcommand\algorithmicensure {\textbf{Output:} }
    \caption{alternating optimization for WD-PSL}
    \label{alg:1}
    \begin{algorithmic}[1]
        \REQUIRE $\left ( x^{i},\textbf{y}^{i} \right)_{i=1}^{n}$, $\epsilon$

        \ENSURE optimized $\omega$

        \label{alg:1}
        \STATE Initialize $P_{0}$, $t = 0$
        \REPEAT  \STATE Optimize $\omega_{t}$ by solving the
minimization problem defined in Equation (17); \STATE  Compute $P_{t}$ according to Equation (22);\STATE $t = t + 1$;
        \UNTIL{$\left | \left
( J(\omega _{t}) - J(\omega _{t-1}) \right )\slash J(\omega _{t})
\right |< \epsilon $}
    \end{algorithmic}
\end{algorithm}

To better guide iterative learning, we initialize $P$ by exploring K-nearest neighborhoods’ annotations. Generally, the instances with same or similar features are more likely to have the same label. Extending this rule to the K--nearest instances with multiple ambiguous annotations, the confidence value for each candidate can be slightly differentiated to effectively exploit the ground-truth label for the subsequent learning process. For the $j_{th}$ candidate of $i_{th}$ instance, the initial $P_{ij}$ is defined as 
\begin{equation}
P_{(0){ij}}=\frac{k_{ij}}{K_{i}},
\end{equation}
where $K_i$ denotes the specified K-nearest neighborhoods of $i_{th}$ instance, and $k_{ij}$ is the occurrences of $j_{th}$ candidate in the summarized ambiguous annotations of K instances.

When $P$ is fixed, Equation (17) can be easily solved by quadratic programming. In order to further reduce the number of constraints, we employ Cutting Plane algorithm \cite{re28} for efficient optimization.

\subsection{Comparison between partial losses}

The ``average" strategy proposed by Lou and Hampercht \cite{re26} is to generalize the Convex Loss for Partial Labels to partial annotations, which gives
\begin{equation}
\scalebox{0.8}{$
\begin{aligned}
J_{0}(\omega )=\min \limits_{\omega}\frac{1}{2}\left \| \omega
\right \|^{2}+\frac{C_{1}}{n}\sum_{i=1}^{n}\max\left (
0,1-\frac{1}{\left \| \textbf{y}^{i}\right \|}\sum_{{y}'\in
\textbf{y}^{i}}E\left ( x^{i},{y}'\right )\right )\\
+\frac{C_{2}}{n}\sum_{i=1}^{n}\sum_{{y}''\in
\mathcal{Y}\backslash\textbf{y}^{i}}\max\left ( 0,1- E\left(
x^{i},{y}''\right)\right ),
\end{aligned}$}
\end{equation}

Candidate Labels for Local Parts (CLLP) was proposed to address the disambiguation between the ground-truth label and other labels. The objective is decomposed as
\begin{equation}
\scalebox{0.85}{$
\begin{aligned}
J_{1}(\omega )=\min \limits_{\omega}
\frac{C_{1}}{n}\sum_{i=1}^{n}\left [\max\limits_{{y}'\in
\textbf{y}^{i}} \Delta (y_{*}^{i},{y}')+\left \langle \omega,E\left
( x^{i},{y}'\right ) \right \rangle -  \left \langle \omega,E\left (
x^{i},y_{*}^{i}\right ) \right \rangle
\right]\\
+\frac{C_{2}}{n}\sum_{i=1}^{n}\left [\max\limits_{{y}''\in
\mathcal{Y}\backslash\textbf{y}^{i}} \Delta (y_{*}^{i},{y}'')+\left
\langle \omega,E\left ( x^{i},{y}''\right ) \right \rangle -  \left
\langle \omega,E\left (
x^{i},y_{*}^{i}\right ) \right \rangle \right] \\
+ \frac{1}{2}\left \|\omega \right \|^{2},
\end{aligned}$}
\end{equation}
where $y_{*}^{i}$ is the true label of $x^{i}$.

\subsubsection{Upper bound}

We compare the average strategy with our objective by restricting structured output to piecewise setting. Generally, the ``average" strategy treats all candidate labels as true labels. The equivalent formulation is expressed as

\begin{equation}
\scalebox{0.8}{$
\begin{aligned}
J_{0}(\omega )=\min \limits_{\omega} \frac{C}{n}\sum_{i=1}^{n}\left
[\max\limits_{y\in \mathcal{Y}\backslash\textbf{y}^{i}}\Delta
(\textbf{y}^{i},y)+\left \langle \omega,E\left ( x^{i},y\right )
\right \rangle -  \left \langle \omega,E\left (
x^{i},\textbf{y}^{i}\right ) \right \rangle
\right] \\
+ \frac{1}{2}\left \| \omega \right \|^{2},
\end{aligned}$}
\end{equation}

$\Delta (\textbf{y}^{i},y)$ is defined as
\begin{equation}
\Delta (\textbf{y}^{i},y)= \left\{
\begin{array}{lcl}
s & & y\notin
\textbf{y}^{i},\\
0  & & y\in \textbf{y}^{i},
\end{array} \right.
\end{equation}
where $s$ is the number of candidate labels in $\textbf{y}^{i}$.

Based on the analysis of $J_{0}(\omega )$ and $J(\omega) $, we obtain the following inequality:
\begin{equation}
J_{0}(\omega)< J(\omega),
\end{equation}
which means $J(\omega)$ upper bounds $J_{0}(\omega)$.

\subsubsection{Efficiency}

Given $N$ sequences, the average length of sequences is $L$. Here we divide the long sequence into pieces with a single transition factor, which generate $N*(L-1)$ pieces. The number of candidate labels and all possible labels for each element in sequences is set to $k$ and $\mathcal{Y}$ respectively. In the framework of soft-margin formulation, the above three partial losses specify different number of constraints which are expressed as follows:
\begin{equation}
\left\{
\begin{array}{lcl}
J_{0}(\omega): N*\left ( \mathcal{Y}^{L}-k^{L}+1 \right ),\\
J_{1}(\omega): N* \left (\mathcal{Y}^{L}-1 \right ),\\
J(\omega): N*(L-1)*\mathcal{Y}^2.
\end{array} \right.
\end{equation}

In most cases the average length of sequences is larger than $10$. The optimization for $J_{0}(\omega)$ and $J_{1}(\omega)$ has to handle large number of constraints. Piecewise training in the proposed objective greatly reduce the number of constraints, which can improve the efficiency of parameter learning as a result.

\section{Experiments}
We perform experiments on two NLP tasks: POS tagging and Chunking. Exact annotation for these two tasks is not feasible because of words' ambiguity. For example, ``like" can be a noun or adverb, then it can be in the noun phrase chunk or just the first of word of a chunk of adverb. Applying partial structured output learning to POS tagging and Chunking effectively handle ambiguous label annotations.

\subsection{Data Sets}

POS tagging: This task is to annotate each word in a sentence a particular part of speech (e.g., determiner and adjective). We choose Penn Treebank \cite{re29} corpus with the widely used data set Wall Street Journal (WSJ). For original dataset, sections 15-18 as training data (211727 tokens) and section 20 as test data (47377 tokens). 

Chunking: This task aims to divide the sentence into groups such as noun phrases and verb groups. We use the same dataset in the shared task of Chunking in CoNLL 2000 \cite{re30}.

Table \uppercase\expandafter{\romannumeral1} summarizes the details of two datasets which include the number of sequences for training (Train*) and testing (Test*), and the number of class labels (Labels*). To avoid the formulation of very large quadratic programming optimization, we construct a small scale training data from sections 15-18, which can preserve the variety of labels. The test data is kept the same with benchmark in the experiment. 

Each sequence in the training set is divided into pieces and each piece is treated as an individual sample. We provide candidate structured outputs for each sample by random generation. This strategy randomly generates the fixed size of candidate structured outputs for each sample and guarantees that the ground-truth label is among candidates. As shown in Table \uppercase\expandafter{\romannumeral1} , $cl=\left \{ 2,3,4 \right \}$ is the number of candidate structured outputs. $p=\left \{ 0.1,0.3,...,0.9 \right \}$ is the proportion of annotated training samples.

\begin{table}
\centering
\caption{Summarization of POS tagging of Chunking datasets}
\begin{tabular}{lccc|l|} 
\hline
\multicolumn{1}{c}{Task} & Train* & Test* & Labels* & \multicolumn{1}{c}{Setting}                                                                                                           \\ 
\hline
POS tagging              & 800    & 2012  & 43      & \multirow{2}{*}{\begin{tabular}[c]{@{}l@{}} $cl=\left \{ 2,3,4 \right \}$\\$p=\left \{ 0.1,0.3,...,0.9 \right \}$ \end{tabular}}  \\
Chunking                 & 1500   & 2012  & 26      &                                                                                                                                       \\
\hline
\end{tabular}
\end{table}

\subsection{Baseline Methods}
We compare the proposed weak disambiguation with four commonly employed disambiguation strategy: random disambiguation, average disambiguation (AD), identification disambiguation (ID) and disambiguation-free. Representative approaches are stated as follows:

S-SVM \cite{re1}: a fully supervised method for structured output learning.

NAIVE \cite{re27}: a S-SVM based method but randomly chooses a label from candidate labels as the true label.

CLPL \cite{re26}: AD-based method for disambiguating the candidate labels and non-candidate labels.

PL-SVM \cite{re19}: ID-based method which aims to maximum the margin between the ground-truth label and the best prediction of wrong label.

CLLP \cite{re27}: ID-based method which incorporates two types of constraints. The first is to maximum the margin between the ground-truth label and other candidate labels while the other is for disambiguating the ground-truth label with non-candidate labels.

IPAL \cite{re31}: ID-based method which identifies the ground-truth label via an iterative label propagation procedure [ suggested setup: k = 10].

PALOC \cite{re32}: ID-based method which induces the multi-class classifiers with one-vs-one decomposition strategy by considering the relevancy of each label pair in the candidate label set.

PL-ECOC \cite{re22}: a disambiguation-free strategy by applying error-correcting output codes (ECOC) to partially labeled instances [suggested setup: the codeword length $L =\left \lceil 10\cdot \log_{2} (q)\right \rceil$.

Considering the large number of constraints specified in the PLL-based structured output learning methods (e.g. CLPL and CLLP), we restrict these models to piecewise setting. For all of the above max-margin based methods, we prepare the validation dataset with half the size of training set and adopt the grid search method to select $C$ parameter from grids [0.01,0.1,1,10,100].

\subsection{Experimental Results}

We first vary $cl$ from 2 to 4 and $p$ from 0.1 to 0.9, and measure the performance of POS tagging and Chunking with F1 score which has been widely used in NLP tasks. 

1) POS tagging. The results on POS tagging test dataset are reported in Figure 6. We made the following observations:

\begin{itemize}

\item
The proposed WD-PSL always outperforms the other partial label learning based methods (e.g., PL-SVM, CLLP and PL-ECOC). Compared with S-SVM, the fully supervised algorithm, WD-PSL achieves better performance with appropriate setting of $p$.

\item 
By increasing the proportion of annotated training instances, the performance of WD-PSL, CLPL  and PL-SVM are more stable than the other baselines. Furthermore, there is no significant positive (or negative) relationship between the performance and the proportion of annotated training instances for most of partial label learning based methods.

\item
The performance of the baselines and the proposed WD-PSL do not vary significantly as the number of candidate labels increases. For example, the average F1 score of WD-PSL for $cl=2$ is 69.65\% and the average F1 score for $cl=4$ is 68.60\%.

\end{itemize}

\begin{figure*}[htb]
\minipage{0.32\textwidth}
  \includegraphics[width=\linewidth]{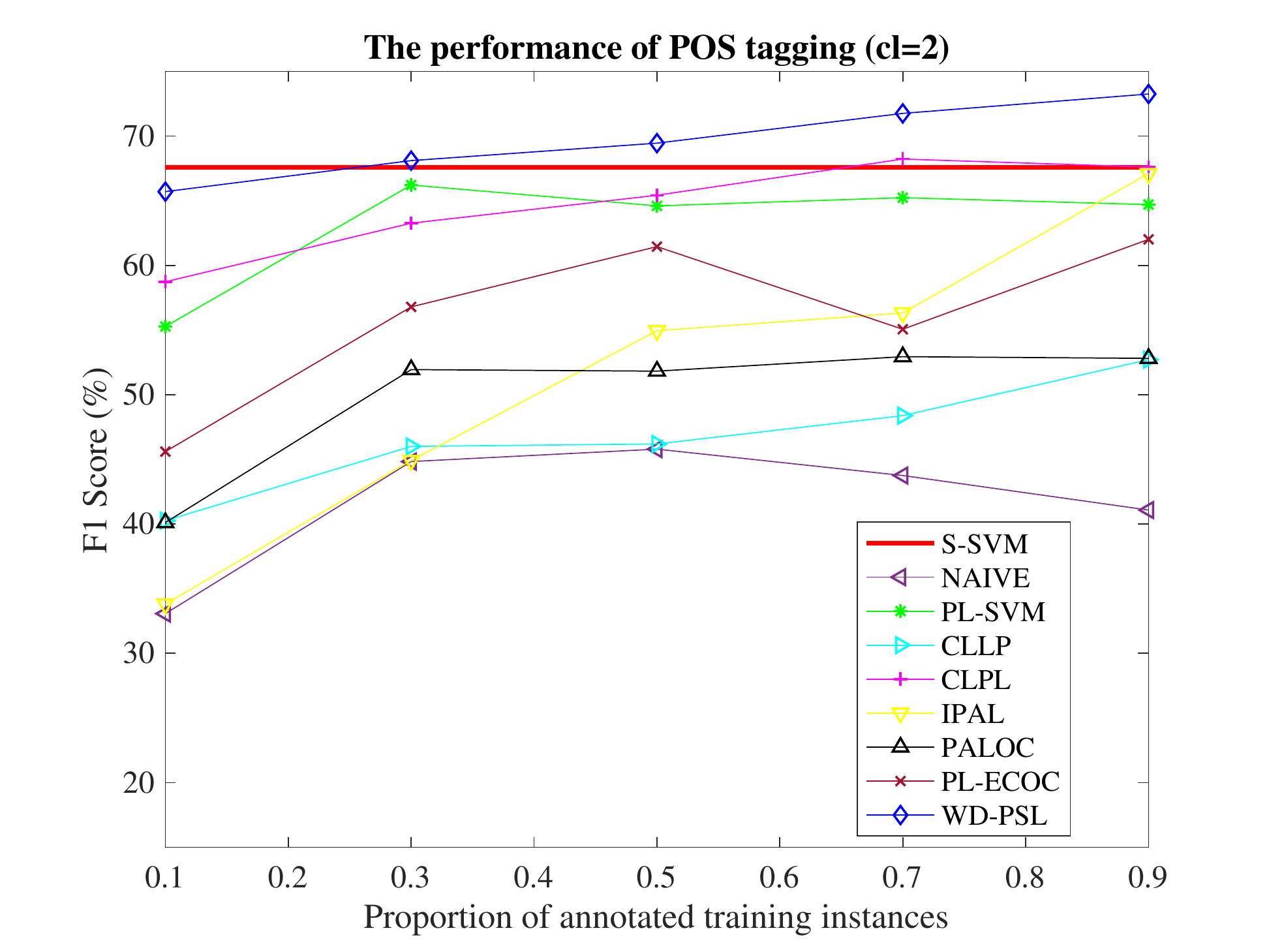}
  \subcaption{$cl$=2)}\label{fig:awesome_image1}
\endminipage\hfill
\minipage{0.32\textwidth}
  \includegraphics[width=\linewidth]{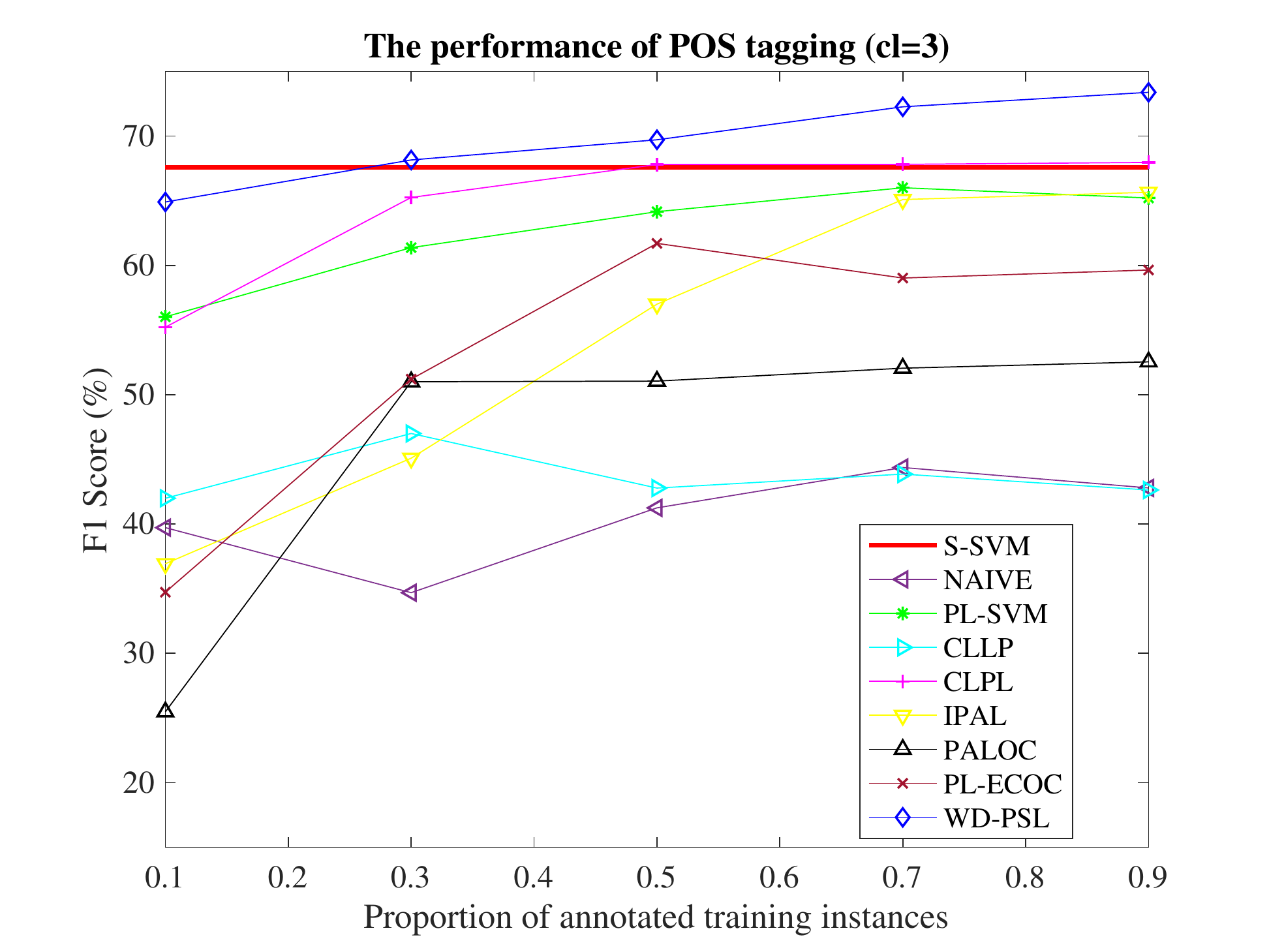}
  \subcaption{$cl$=3}\label{fig:awesome_image2}
\endminipage\hfill
\minipage{0.28\textwidth}%
  \includegraphics[width=\linewidth]{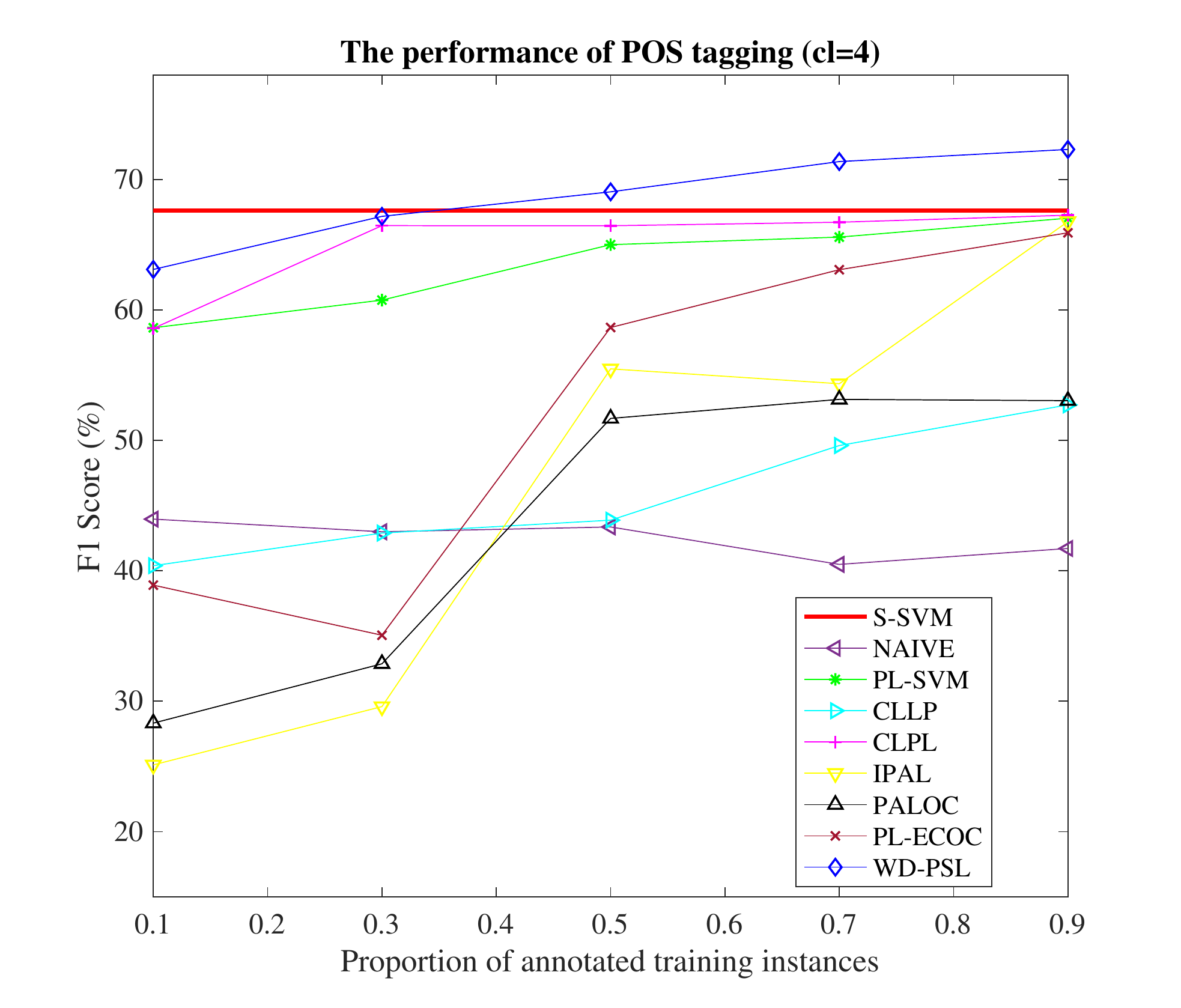}
  \subcaption{$cl$=4}\label{fig:awesome_image3}
\endminipage
\caption{The performance of POS tagging.}
\end{figure*}

2) Chunking. The results on Chunking are reported in Figure 7. The following observations can be made:

\begin{itemize}

\item 
As shown in Figure 9, 10 and 11, WD-PSL outperforms the other partial label learning based methods in most cases. In the task of Chunking, WD-PSL always achieves better performance than S-SVM.

\item 
By increasing the proportion of annotated training instances, the performance of WD-PSL, CLPL and PL-SVM are more stable than the other baselines. Similar to the task of POS tagging, there is no significant positive (or negative) relationship between the performance and the proportion of partially annotated sequences.

\item
In most cases, the performance of WD-PSL and other partial label learning methods does not change significantly as the number of candidate labels increases.

\end{itemize}

\begin{figure*}[htb]
\minipage{0.32\textwidth}
  \includegraphics[width=\linewidth]{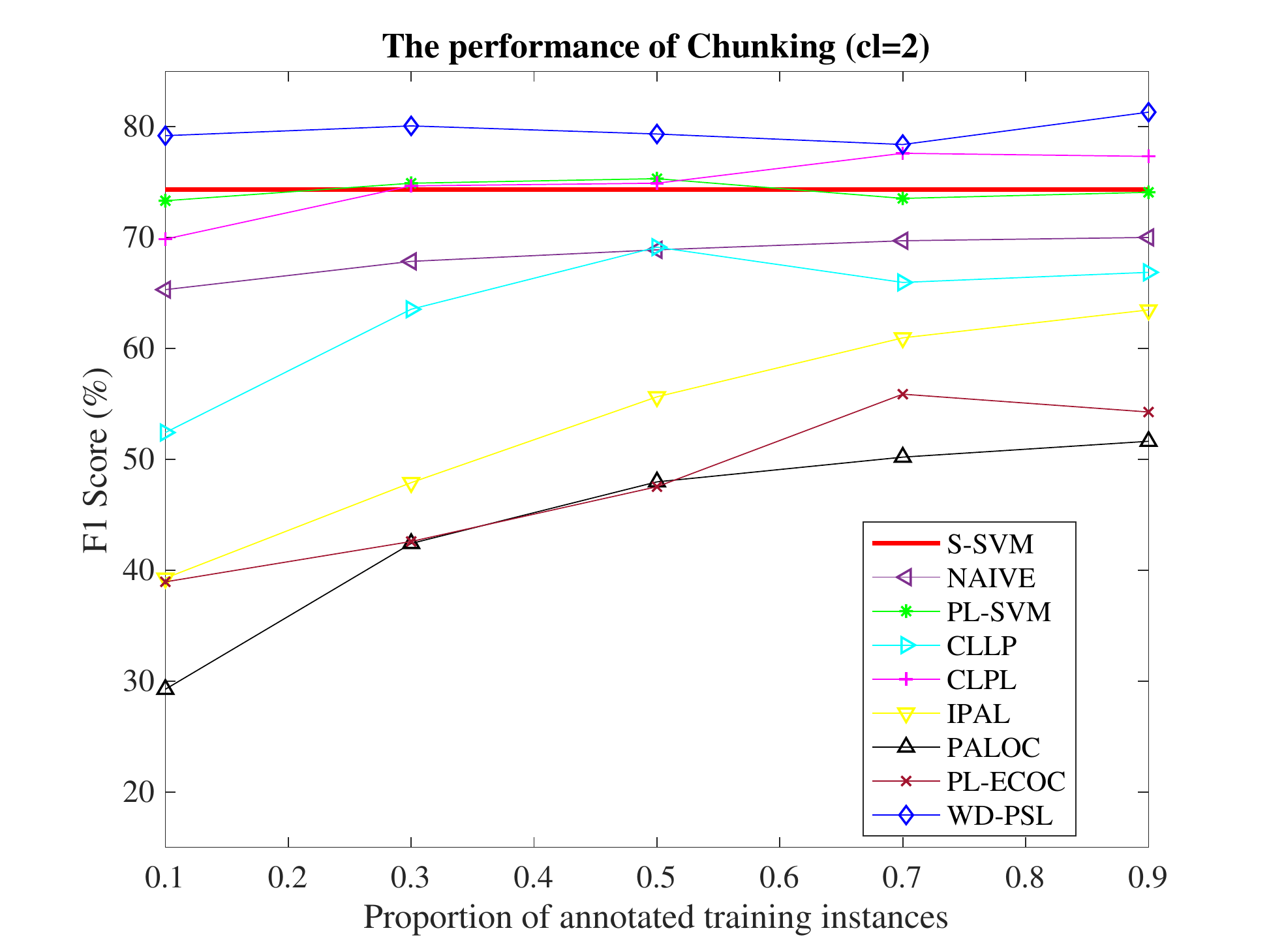}
  \subcaption{$cl$=2)}\label{fig:awesome_image1}
\endminipage\hfill
\minipage{0.32\textwidth}
  \includegraphics[width=\linewidth]{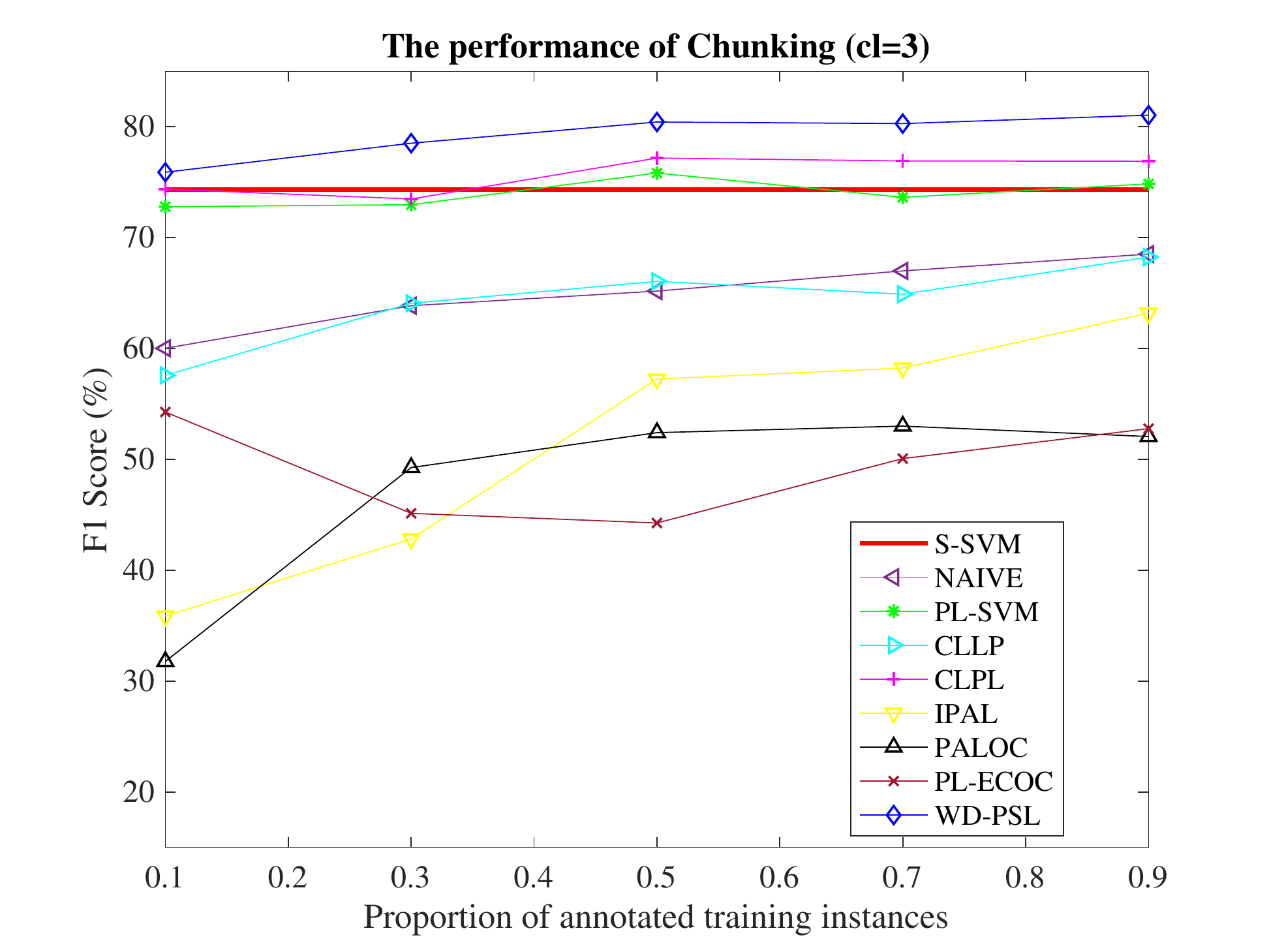}
  \subcaption{$cl$=3}\label{fig:awesome_image2}
\endminipage\hfill
\minipage{0.32\textwidth}%
  \includegraphics[width=\linewidth]{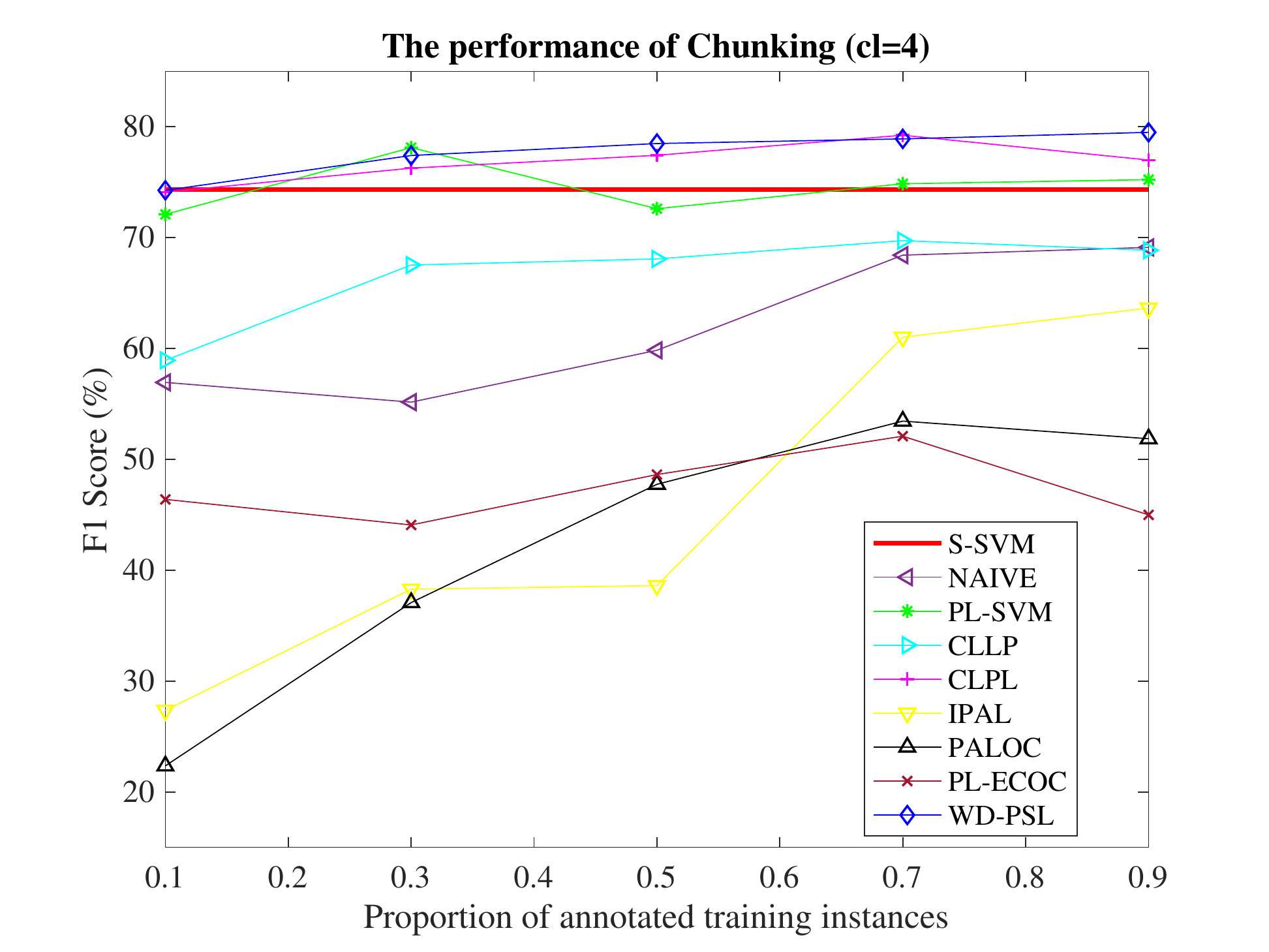}
  \subcaption{$cl$=4}\label{fig:awesome_image3}
\endminipage
\caption{The performance of Chunking.}
\end{figure*}

Considering the vocabulary size of whole dataset, we further employ the training set and record the performance of each comparing methods with 3$\times$ 5-fold cross validation, which repeats the whole cross validation 3 times. Based on the observations from Figure 6 to 11, the comparing methods do not vary significantly for different number of candidate labels and most of comparing methods performs stably with above-average results in the setting of $p=0.5$, therefore we choose $cl=3$ and $p=0.5$.  Table  \uppercase\expandafter{\romannumeral2} presents the results of each PLL-based methods on POS tagging and Chunking. Meanwhile the results of one-tailed t-test at 5\% significance level are recorded. As shown in Table  \uppercase\expandafter{\romannumeral2}, we can observe that WD-PSL is superior to the other comparing methods on Chunking, and for the task of POS tagging, WD-PSL is comparable to CLPL and outperforms the other comparing algorithms.

\begin{table}
\centering

\caption{F1 score (mean$\pm$std) of each PLL-based methods on POS tagging and Chunking.}

\begin{threeparttable}

\begin{tabular}{lll}
\hline
        & POS tagging(\%)                & Chunking(\%)                    \\
\hline
WD-PSL  & $72.05\pm1.13$ & $75.26\pm1.47$  \\
PLSVM   & $65.25\pm1.88$$\bullet $  & $72.21\pm3.66$$\bullet $   \\
CLPL    & $67.68\pm2.99$$\circ  $& $72.55\pm2.49$$\bullet $    \\
CLLP    & $42.37\pm1.17$$\bullet $ & $65.17\pm3.83$$\bullet $  \\
NAIVE   & $40.68\pm3.27$$\bullet $  & $70.75\pm4.21$$\bullet $  \\
IPAL    & $58.40\pm4.87$$\bullet $ & $59.22\pm3.45$$\bullet $  \\
PALOC   & $43.45\pm2.17$$\bullet $  & $36.72\pm3.80$$\bullet $  \\
PL-ECOC & $44.70\pm2.59$$\bullet $ & $40.99\pm2.67$$\bullet $  \\
\hline
\end{tabular}
\begin{tablenotes}
 \item[] $\bullet $/ $\circ  $ denotes whether the performance of WD-PSL is statistically superior/inferior to the comparing methods (one tailed t-test at 5\% significance level).
 \end{tablenotes}
 \end{threeparttable}
\end{table}

Based on the analysis of the above results obtained from POS tagging and Chunking tasks, in some cases the proposed WD-PSL outperforms CLPL by a narrow margin. CLPL only considers the equal contribution of all candidates while WD-PSL assigns different confidence value to the label in candidates. When handling the candidates that are very similar (e.g. (NN,NNS)), the performance of WD-PSL and CLPL can be very close. However, WD-PSL further addresses the importance of the ground-truth label with weak disambiguation strategy, which guarantees the stable performance in different configuration of candidate labels. Furthermore, it is worth noting that WD-PSL can outperform S-SVM with appropriate proportion of exact annotation. Since the quality and quantity of training data affect the learning, S-SVM cannot handle noisy training set while WD-PSL can improve the performance by learning from ambiguous labels.

\subsection{Sensitivity Analysis}

In this section, we study the influence of two regularization parameter $C_{1}$ and $C_{2}$ on the performance of WD-PSL. By varying $C_1 (C_2)$ from 0.01 to 100, the experiment is conducted on POS tagging and Chunking with fixed $cl = 3$ and $p = 0.5$. Figure 12 and 13 report the performance of the performance of WD-PSL under different configurations. The sensitivity analysis is made as follows:

\begin{itemize}

\item 
As shown in Figure 8 and 9, generally WD-PSL achieves better performance on POS tagging and Chunking with larger value of $C_{2}$. Continuing to decrease the value of $C_{2}$ may degrade the performance. Furthermore, smaller value of $C_1$ is more favorable in the Chunking. Both POS tagging and Chunking obtain the best performance with smaller $C_1$.

\item 
The objective function of WD-PSL aims to maxmize the margin between the candidates and non-candidates and address the different contribution of the label in candidates. Although the performance does not change significantly by varying $C$ parameter, setting a smaller value of $C_{2}$ allows the label with lower confidence value in candidates to be classified as non-candidates, which can improve the discriminative ability of the model.

\end{itemize}

\begin{figure}
\centering
\includegraphics[width=3.7in,height = 2.5in ]{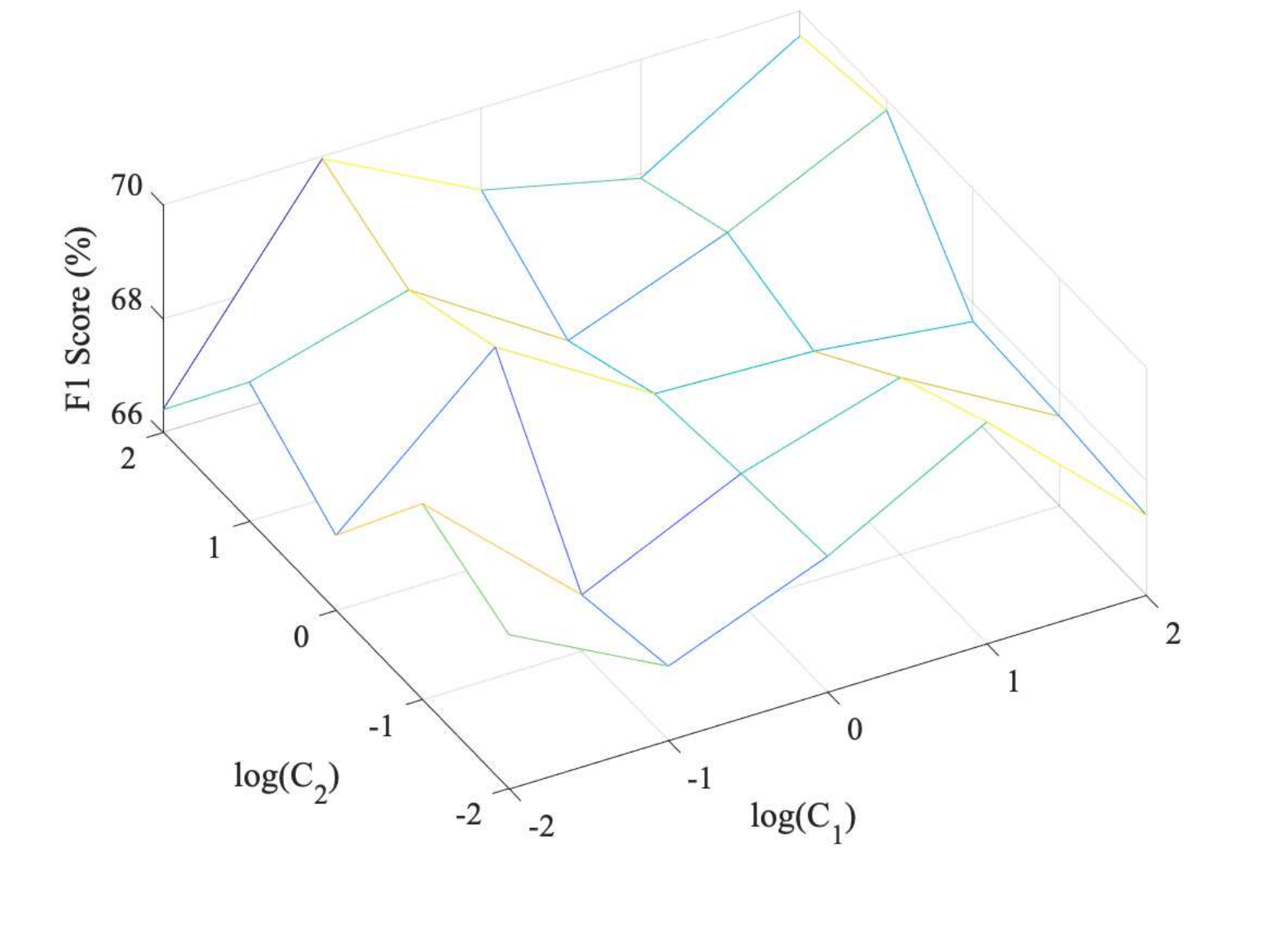}
\caption {Impact of $C$  parameters on the performance of POS tagging.}
\label{fig:secondfigure}
\end{figure}

\begin{figure}
\centering
\includegraphics[width=3.7in,height = 2.5in ]{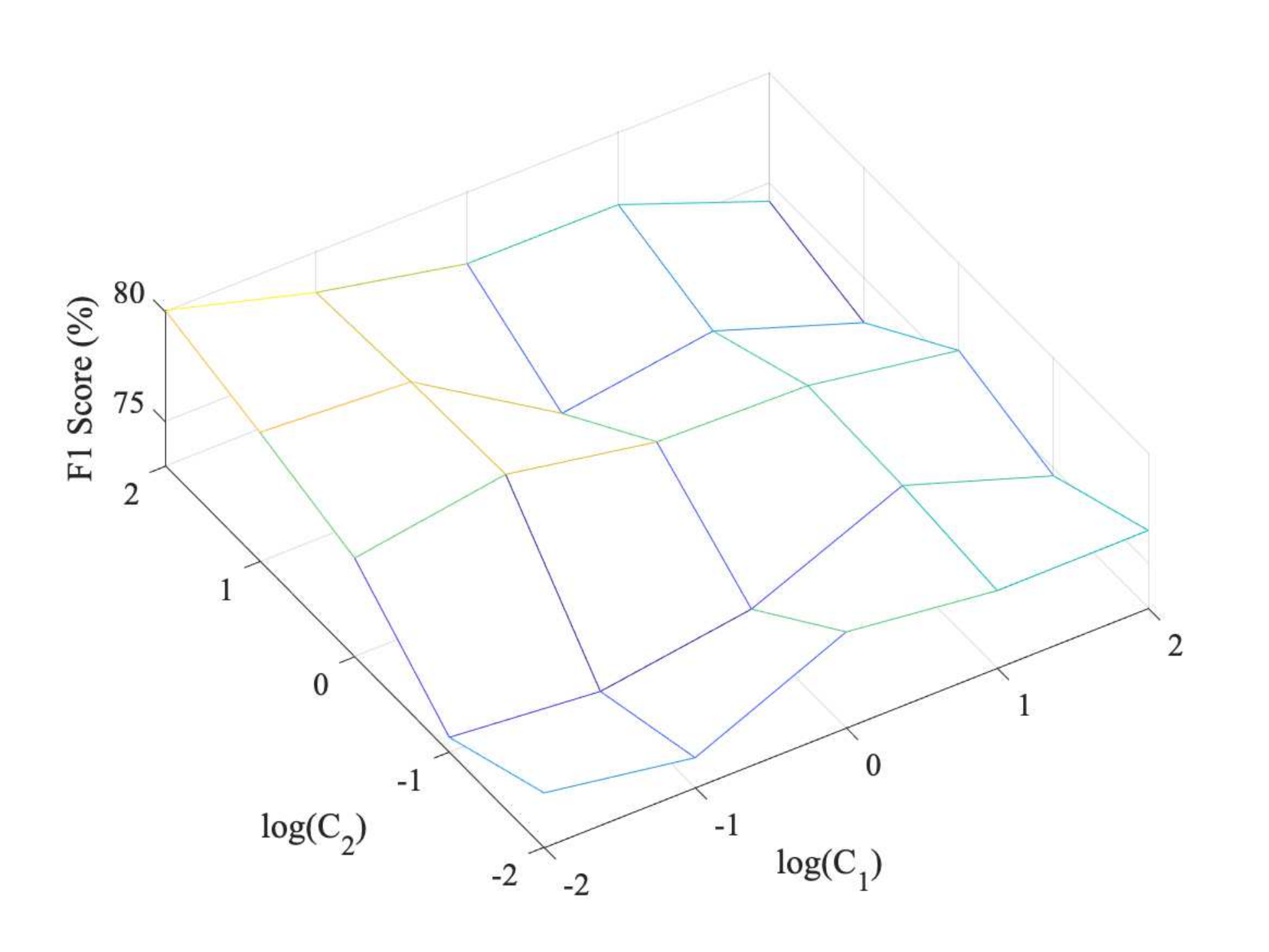}
\caption {Impact of $C$ parameters on the performance of Chunking.}
\label{fig:secondfigure}
\end{figure}

\subsection {Weak disambiguation strategy analysis}

The proposed weak disambiguation strategy measures the probability of the candidate being the ground-truth by assigning the different confidence value. In this section, we study the effect of confidence value assigned by weak disambiguation strategy. 

First, we select some training samples from POS tagging and Chunking to present the changes of confidence value for each candidate in alternating optimization. For these two tasks, we set $cl=3$ and $p=5$.

We choose candidate structured outputs (`VBZ', `JJ', `NNS'), (`NN', ```', `NNP'), (`NN', `IN', `DT') in the task of POS tagging. The ground-truth is (`NN', `IN', `DT'). Figure 10 (a) presents the change of confidence value for each candidate in alternating optimization. We can see that (`NN', `IN', `DT') can be easily identified from the candidates. And the candidate (`VBZ', `JJ', `NNS') always obtains the minimum score in the iteration. The contribution of the ground-truth (`NN', `IN', `DT') is much greater than other candidates, which obviously benefits parameter learning.

Then we choose candidate structured outputs (`O', `B-ADVP', `O', `B-VP'), (`I-SBAR', `B-NP', `B-VP', `I-VP'), (`O', `B-NP', `I-NP', `I-NP') in the task of Chunking. The ground-truth is (`O', `B-NP', `I-NP', `I-NP'). Figure 10 (b) presents the change of confidence value fo each candidate in alternating optimization. Although the ground-truth (`O', `B-NP', `I-NP', `I-NP') does not always obtain the maximum score during itertaitons, the weak disambiguation strategy enables (`O', `B-NP', `I-NP', `I-NP') to make more or less contributions in each iteration, which can reduce the negative effects of wrong ground-truth label assignment.

Second, we compare the performance with different confidence setting. By considering the equal contribution of each candidate, the confidence value of each candidate is assigned with $1/cl$. In the experiment, the weak disambiguation strategy is denoted as WD while the equal confidence setting is denoted as AVG. Figure 10 (c) shows the performance with WD and AVG for POS tagging (POS) and Chunking (CK). It can be observed that the proposed WD always outperforms AVG setting. 

Generally, weak disambiguation strategy addresses different contribution of the label in candidates and thus greatly reduces the negative effects of wrong ground-truth label assignment in iterative optimization, which narrows the gap between average disambiguation and unique disambiguation strategy.

\begin{figure*}[!htb]
\minipage{0.32\textwidth}
  \includegraphics[width=\linewidth]{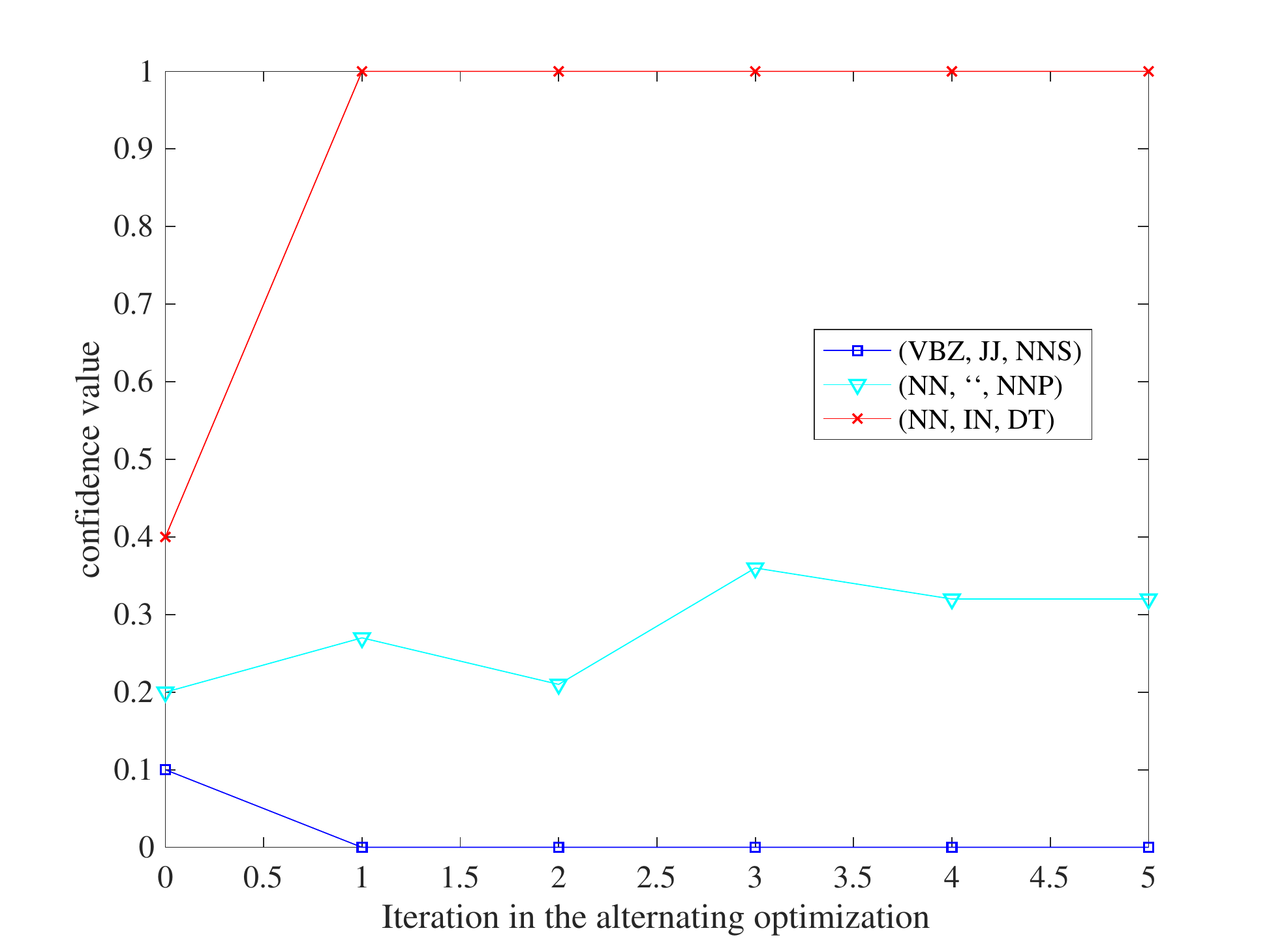}
  \subcaption{}\label{fig:awesome_image1}
\endminipage\hfill
\minipage{0.32\textwidth}
  \includegraphics[width=\linewidth]{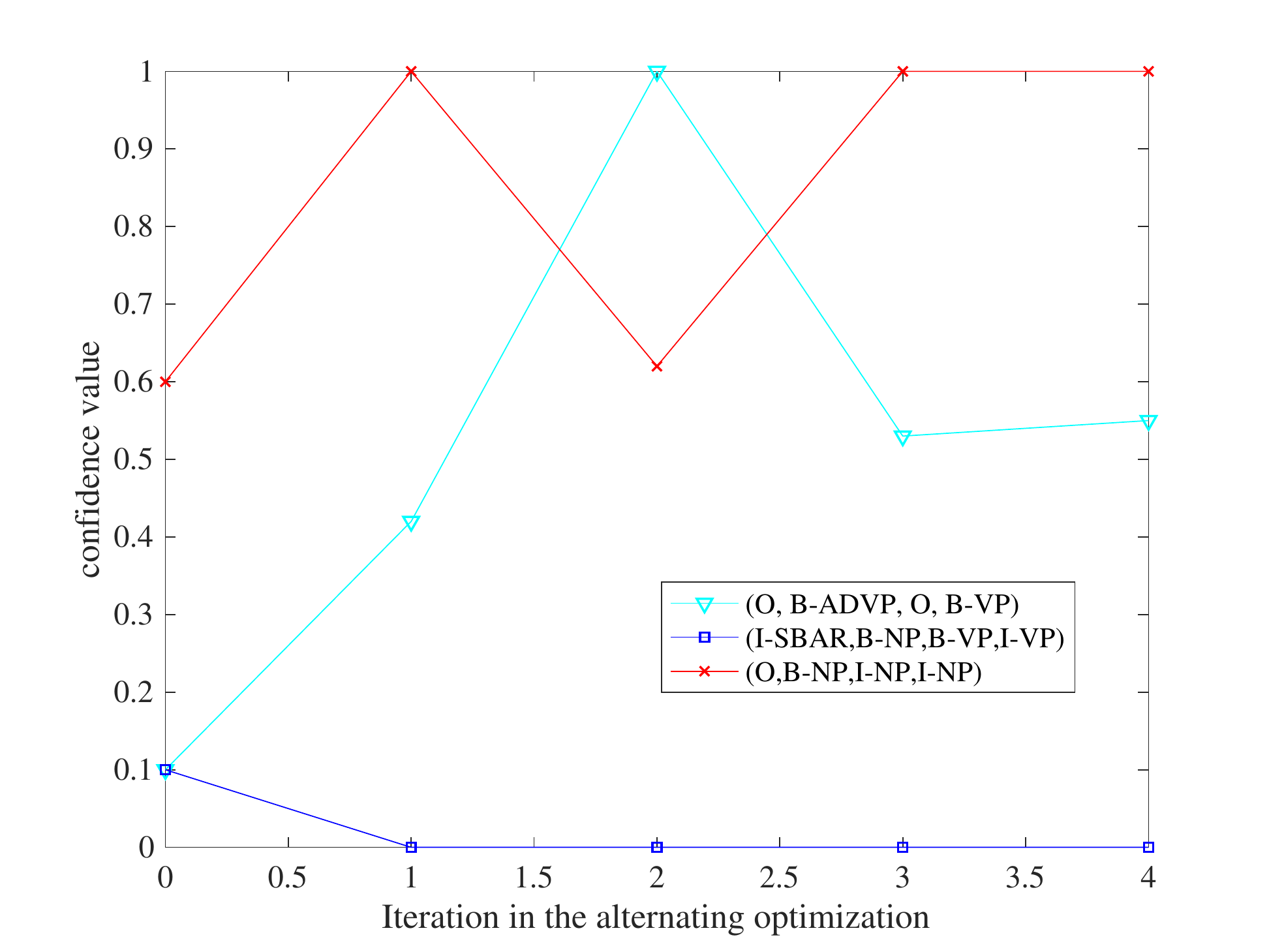}
  \subcaption{}\label{fig:awesome_image2}
\endminipage\hfill
\minipage{0.32\textwidth}%
  \includegraphics[width=\linewidth]{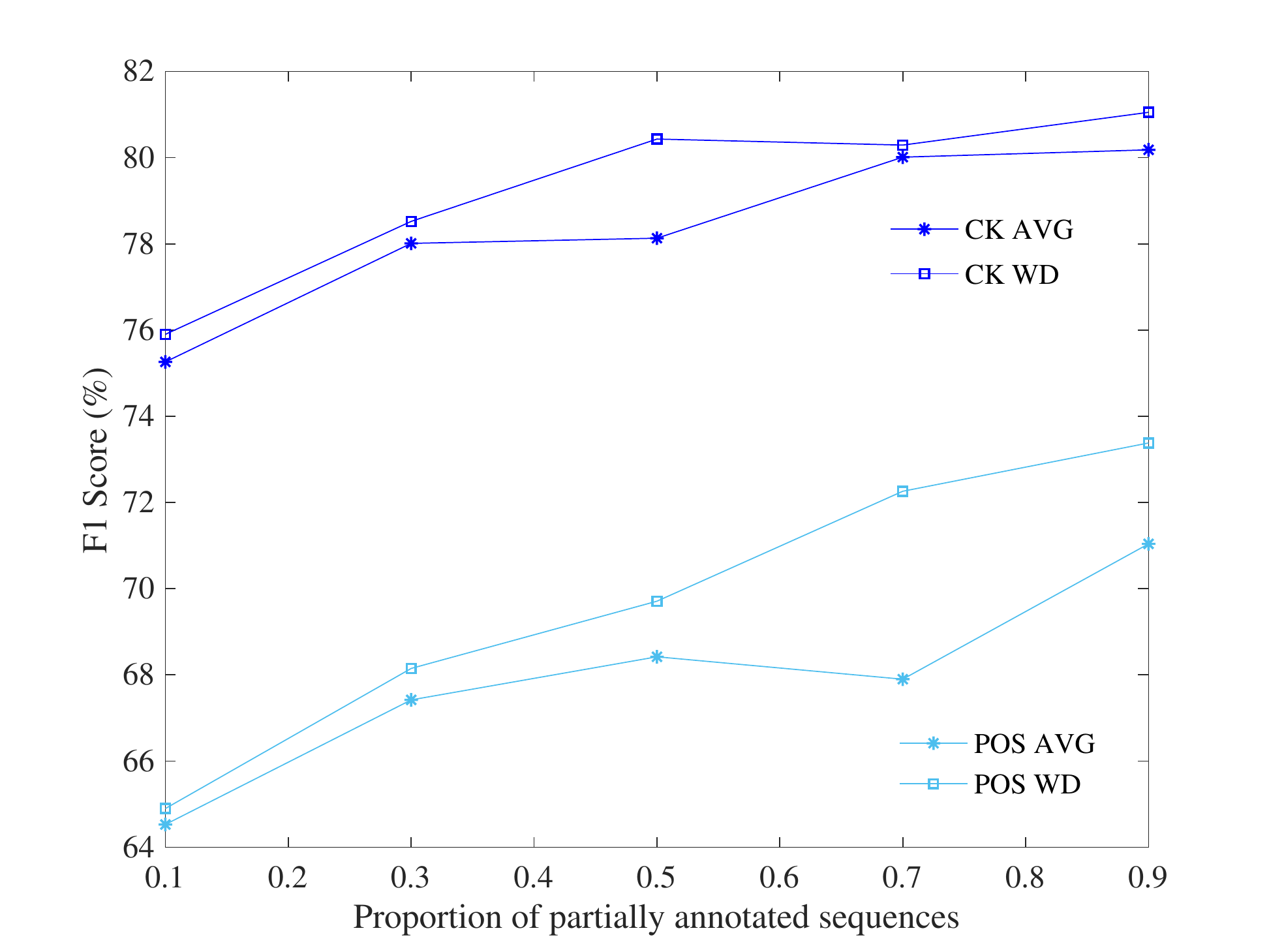}
  \subcaption{}\label{fig:awesome_image3}
\endminipage
\caption{(a) Confidence changes in alternating optimization for POS tagging. (b) Confidence changes in alternating optimization for Chunking. (c) The performance of POS tagging and Chunking (cl=3).}
\end{figure*}

\section{Conclusion}
In this paper, we propose a novel model WD-PSL to realize weak disambiguation for partial structured output learning. The proposed WD-PSL generalizes the piecewise learning to partial label learning which avoids handling large number of candidate structured outputs. Furthermore, the ``weak disambiguation" strategy is incorporated into the two large margins formulation, which greatly reduce the negative effects of wrong ground-truth label assignment in the learning process and thus improve the performance of structure prediction. We conducted the experiments on the tasks of POS tagging and Chunking. The experimental results show that the proposed WD-PSL can outperform the baselines. Furthermore, WD-PSL is less sensitive to the proportion of annotated training samples, which performs stably with large proportion of ambiguous annotations. In the future work, we will focus on the efficient inference solution for learning from partially annotated sequence.

\section{Acknowledgements}
This work is supported by a project donated by Mr. MW Lau of CityU project No. 9220083.
\bibliographystyle{ieeetr}
\bibliography{ref2}

\end{document}